\documentclass[conference]{IEEEtran}
\IEEEoverridecommandlockouts
\usepackage{amsmath,amssymb,amsfonts}
\usepackage{algorithmic}
\usepackage{graphicx}
\usepackage{textcomp}
\usepackage{xcolor}
\usepackage{multirow}
\usepackage{multicol}
\usepackage{todonotes}
\usepackage{url}
\usepackage{booktabs}
\usepackage[
    backend=biber,
    style=science,
]{biblatex}
\providecommand{\keywords}[1]
{
  \small	
  \textbf{\textit{Keywords---}} #1
}

\begin{document}

\title{Chemical Identification and Indexing in PubMed Articles via BERT and Text-to-Text Approaches\\
}

\author{
Virginia Adams\textsuperscript{\textsection},
Hoo-Chang Shin\textsuperscript{\textsection},
Carol Anderson\textsuperscript{\textsection},
Bo Liu\textsuperscript{\textsection},
Anas Abidin\textsuperscript{\textsection},
\\
NVIDIA / Santa Clara, California, USA\\
\texttt{\{vadams;hshin;carola;boli;aabidin\}@nvidia.com}
}

\maketitle

\begingroup\renewcommand\thefootnote{\textsection}
\footnotetext{In reverse alphabetical order - authors contributed equally.}
\endgroup

\begin{abstract}
The Biocreative VII Track-2 challenge consists of named entity recognition, entity-linking (or entity-normalization), and topic indexing tasks -- with entities and topics limited to chemicals for this challenge. Named entity recognition is a well established problem and we achieve our best performance with BERT-based BioMegatron models. We extend our BERT-based approach to the entity linking task. After second stage pretraining BioBERT with a metric-learning loss strategy called self-alignment pretraining (SAP), we link entities based on the cosine similarity between their SAP-BioBERT word embeddings.  Despite the success of our named entity recognition experiments, we find the chemical indexing task generally more challenging.

In addition to conventional NER methods, we attempt both named entity recognition and entity linking with a novel text-to-text or ``prompt'' based method that uses generative language models such as T5 and GPT.
We achieve encouraging results with this new approach. \\
\end{abstract}

\keywords{\textbf{\textit{
BERT, BioMegatron, GPT, T5, Entity-Linking
}}}

\begin{figure*}[htbp]
\centering
\includegraphics[width=.95\textwidth]{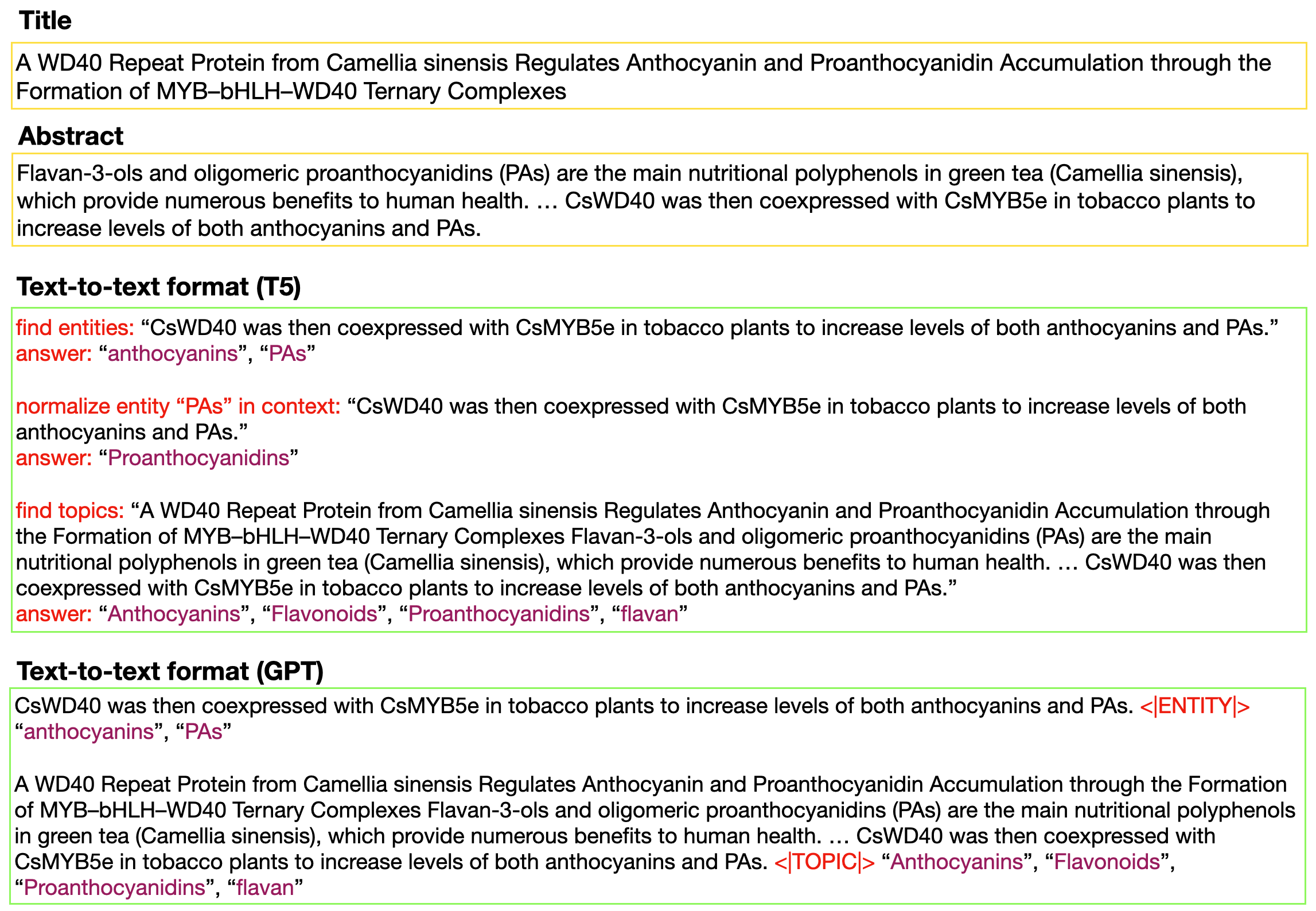}
\caption{Examples of converting NER, entity-linking, and indexing tasks into text-to-text format for T5 and GPT models. First, we break long paragraphs into sentences. Then we convert each sentence into a question and answer for T5 models, and more simple format with special prompting tokens for GPT models.}
\label{fig:text-to-text_preprocessing}
\end{figure*}

\section{Introduction}
The body of scientific literature is vast and rapidly growing, with over 1.5 million paper citations added to PubMed alone last year\footnote{count retrieved from https://pubmed.ncbi.nlm.nih.gov}. In order to make the knowledge in these papers easily discoverable, effective methods are needed to automatically extract entities, link them to standardized concepts in knowledge bases, and index key topics. Track 2 of Biocreative VII~\cite{bc7track2} asks us to develop methods for named-entity recognition (NER), entity linking, and topic indexing of chemical names and topics found in full-text PubMed articles.

NER is a critical first step in extracting information. Current state-of-the-art NER methods use BERT-based models \cite{devlin2018bert,lee2020biobert}. Previous work has shown that using domain-specific vocabularies and pretraining BERT on in-domain text significantly boosts performance for biomedical natural language processing (NLP) tasks \cite{beltagy2019scibert,gu2020domain}. Moreover, Shin et al. \cite{shin2020biomegatron} reported that, compared to other NLP tasks, NER is particularly sensitive to differences in model vocabulary. The beneficial effect of domain-specific vocabulary is likely due to the fact that it is helpful to represent entities as single tokens, whereas general-domain vocabularies tend to break up biomedical terms into multiple subtokens. Experiments by Shin et al. \cite{shin2020biomegatron} also demonstrated that larger models outperform smaller ones in biomedical NLP. 

Following NER, entity linking is another key step in information extraction pipelines. Entity linking is the process of matching concepts mentioned in natural language to their unique IDs and canonical forms stored in a knowledge base. Traditional approaches to entity linking involve string matching, calculating edit distance, and other heuristic based methods. Modern deep learning entity linking strategies generally combine an NER model, candidate generation model, and entity ranking model in a multi-step pipeline \cite{sevgili2021neural}. 

For our Track-2 submission, we use an implementation of Liu et al. \cite{liu-etal-2021-self}'s self-alignment pre-training technique. Self-alignment pre-training is a neural entity linking approach in which a pre-trained language model undergoes a second stage of pre-training and the resulting embedding space is used to assess semantic similarity between two entities.

Recent work has shown that many NLP tasks can be framed as "text-to-text" or "prompt-based" tasks using generative language models. This approach offers the advantage that a single model can be used for different tasks without a need for task-specific layers. Such an approach seems particularly promising for this track since it consists of three different but related sub-tasks. Although this approach has successfully been applied to a variety of NLP problems such as text classification, natural language inference, and summarization, to our knowledge there are no studies to date that attempt to formulate NER, entity linking, or topic indexing as text-to-text tasks. We try this approach by reframing the different tasks as text-to-text problems (Fig. \ref{fig:text-to-text_preprocessing}) and using T5 \cite{raffel2019exploring} and GPT \cite{radford2019language} models fine-tuned on the challenge data to generate "answers" conditioned on "prompts." Our experiments with this novel approach do not achieve the same performance as more established approaches, yet they show some promise and point towards intriguing possible lines of investigation for future research. 

We find that the chemical indexing task is more challenging than NER and entity linking, especially using BERT-based models, despite having successful BERT-based NER models.
We considered developing hierarchical models to map chemical named entities into common ancestors in the MeSH database for topic indexing. However, we find that there is limited overlap between chemical entities found in an article and the chemical topics indexed for that article. Many of the indexed chemical topics do not correspond directly to chemical entities named in the article, and conversely, the chemical entities discussed in an article are not necessarily indexed as topics. With the limited number of labeled articles provided, we found it hard to train models to learn principles of topic indexing.

\section{Dataset}
Annotated training, development, and test data are provided as part of the BioCreative VII Track 2. The data includes 150 full-text PubMed articles. Each article has labeled chemical entities and chemical indexing topics. The chemical entities and topics are provided with their standardized Medical Subject Heading (MeSH) IDs ~\cite{lipscomb2000medical}.
There is also a held-out test set where the gold-standard annotation is not provided - but used for official submission and evaluation. For more details about the dataset please refer to the dataset paper by Islamaj et. al \cite{bc7track2data}.
We report the evaluation scores on the test set with annotations throughout the paper.
The official scores from held-out test set on NER and entity-linking are shown at the end.

We exclusively used the provided training data for the BERT-based experiments.
The text-to-text based methods required large-scale data in order to achieve comparable results to the BERT-based methods. We utilized the BioCreative VII Track 1 large-scale sub-track data ~\cite{drugprot} as an additional training data source for the NER task.
For our GPT-based experiments we also made use of the previous BioCreative V challenge ~\cite{leitner2010overview} data -- CDR and CHEMDNER~\cite{krallinger2015chemdner}.

We used about 15\% random subset of PubMed articles with chemical indexing as supplementary training data for topic indexing - a subset was used due to time constraint.

\section{BERT-based Models}
We employ the widely adopted method of BERT-based token classification for NER.
After verifying the BioBERT~\cite{lee2020biobert} baseline, we experiment with BioBERT-large and BioMegatron~\cite{shin2020biomegatron} models.
For the NER task, training, dev, and test sets with ground-truth annotations are provided, and the prediction on an additional, larger test set is submitted for final official evaluation.

\begin{table}[htbp]
\caption{\label{tab:bert-based-models-perf}Strict and approximate F-1 scores for chemical named entity recognition using BERT-based models on the test set.}
\centering
\resizebox{1\columnwidth}{!}{
\begin{tabular}{lclccc}
\hline
 \textbf{Model} & \textbf{\#Parameters} & \textbf{Vocabulary} & \textbf{Strict.F1}&\textbf{Approx.F1} \\ \hline
 BioBERT-base   & 110m & BERT-uncased & 0.8017 & 0.8998 \\
 BioBERT-large  & 345m & BERT-uncased & 0.7908 & 0.8870 \\
 \textbf{BioMegatron}  & \textbf{345m} & \textbf{bio-cased-50k}   & \textbf{0.8244} & \textbf{0.9274} \\
 BioMegatron    & 345m & bio-uncased-50k & 0.8176 & 0.9256 \\
 BioMegatron    & 345m & bio-cased-30k   & 0.8200 & 0.9183 \\
 BioMegatron    & 345m & bio-uncased-30k & 0.8029 & 0.9139 \\
 BioMegatron    & 800m & BERT-cased      & 0.8114 & 0.9179 \\
 BioMegatron    & 1.2b & BERT-uncased    & 0.8181 & 0.9183 \\
\hline
 \textbf{Ensemble}       & ---  & ---             & \textbf{0.8324} & \textbf{0.9295} \\
\hline
\end{tabular}}
\end{table}

Evaluation scores on the annotated test set are shown in Table~\ref{tab:bert-based-models-perf}.
The best scores are achieved with a BioMegatron model using an in-domain vocabulary set.
BioMegatron models perform better than BioBERT models in general, confirming that larger model size is usually beneficial. There are differences between our implementations of BioBERT-large and BioBERT-base, most notably in their vocabularies, such that the comparison between these two models is not strictly a size comparison.
Meanwhile, larger BioMegatron models using standard BERT vocabularies perform slightly worse than smaller ones with in-domain vocabulary sets, again attesting that in-domain vocabulary is a key factor for the NER task. Model ensembling is known to help provide additional performance gains by combining multiple models' predictions. In one of our two submissions, we ensemble our top five scoring BioMegatron models yielding the test set ensemble score in \ref{tab:bert-based-models-perf}. Our ensemble predictions are produced from a weighted average of the output probabilities from each model. By combining the results from our models, we note a 1 percent increase in F-1 score. 

\begin{table}[htbp]
\caption{\label{tab:bert-based-entity-linking}Precision, recall, and F1-score of entity-linking on the BERT-based model ensemble predictions on the test set.}
\centering
\resizebox{.7\columnwidth}{!}{
\begin{tabular}{lccc}
\hline
        & \textbf{Prec} & \textbf{Rec}&\textbf{F1} \\ \hline
Strict  & 0.4398        & 0.6383      & 0.5208 \\
Approx. & 0.4617        & 0.7844      & 0.5749 \\
\hline
\end{tabular}}
\end{table}

For entity-linking, we adopt the self-alignment pre-training approach described by Liu et al.  \cite{liu-etal-2021-self}. The main idea behind this approach is to use a metric-learning loss to reshape the initial BioBERT embedding space such that synonyms of the same concept are pulled closer together and unrelated concepts are pushed further apart. The concept embeddings from this reshaped space can then be used to build a knowledge base embedding index. This index stores MeSH IDs mapped to their respective concept embeddings in a format conducive to efficient nearest neighbor search. We link query concepts to their canonical forms in the knowledge base by performing a nearest neighbor search-- matching concept query embeddings to the most similar concept embedding in the knowledge base index. 

We use the UMLS dataset to pre-train BioBERT for entity linking. The training dataset consists of pairs of concept synonyms that map to the same ID. At each training iteration, we only select hard examples present in the mini batch to calculate the loss and update the model weights. In this context, a hard example is an example where an entity is closer to an unrelated concept in the mini batch than it is to the synonym concept it is paired with by some margin. Table~\ref{tab:bert-based-entity-linking} shows the entity linking results on the test set using our BERT-based NER ensemble predictions. We observe reasonable scores even though lower than the median scores of the submissions: strict: 0.6873 / 0.7266 / 0.7325; approx: 0.6520 / 0.8075 / 0.7292. We suspect this difference is due to our approach for this subtask relying exclusively on representations of the extracted entities, without consideration for the whole sentence context.

\section{Text-to-Text Approach}

Text-to-text, or ``prompting'' based methods are attractive, as many different tasks can be expressed as natural language question and answer tasks and predictions can be generated for multiple tasks from a single model. We adopt this approach to convert NER, entity-linking, and indexing task into natural language, text-to-text problems. We experiment with two of the popular generative language models: T5~\cite{raffel2019exploring} and GPT~\cite{radford2019language}, both of similar size to BERT-large with about 345m parameters. The second of our two NER submissions come from predictions made by T5. 

Examples of how we pre-process the data into text-to-text format are shown in Figure~\ref{fig:text-to-text_preprocessing}. Long paragraphs are split into sentences. We formulate NER, entity-linking, and indexing tasks as natural language questions for use by the T5 model. For GPT we use a simpler format in which we separate the ``context'' and answer with a special token that is specific to each task, instead of using a natural language question. We did not experiment with entity-linking using GPT. We fine-tune both GPT and T5 on the NLM-Chem train split given by the task organizers.

\begin{table}[htbp]
\caption{\label{tab:text-to-text-ner}Precision, recall, and F1-score of the NER on the test set using T5 and GPT models.}
\centering
\resizebox{.8\columnwidth}{!}{
\begin{tabular}{llccc}
\hline
Model &         & \textbf{Prec} & \textbf{Rec}&\textbf{F1} \\ \hline
\multirow{2}{*}{T5}     & Strict  & 0.7622        & 0.5863      & 0.6628 \\
                        & Approx. & 0.8638        & 0.6551      & 0.7451 \\
\hline
\multirow{2}{*}{GPT}    & Strict  & 0.6684        & 0.5807      & 0.6215 \\
                        & Approx. & 0.7764        & 0.6394      & 0.7013 \\
\hline
\end{tabular}}
\end{table}

Table~\ref{tab:text-to-text-ner} shows the evaluation results for NER using T5 and GPT models.
The scores are lower than those for BERT-based models but encouraging. Loosing the exact entity spans when converting the task to a text-to-text format is an inherent challenge with our text-to-text approach to NER.
We let the generative language models output entities in the ``context'' sentence and, after the output generation, we have to go back to the original ``context'' text to match the generated tokens with the original text in order to calculate the spans for final evaluation. While the text-to-text approach shows promise in the long term, losing the span information and having to re-find the spans from the generated tokens in the original text via string-matching adds to the complexity of the task.

\begin{table}[htbp]
\caption{\label{tab:t5-entity-linking}Precision, recall, and F1-score of entity-linking on the test set for the T5 model.}
\centering
\resizebox{.8\columnwidth}{!}{
\begin{tabular}{llccc}
\hline
Model &         & \textbf{Prec} & \textbf{Rec}&\textbf{F1} \\ \hline
\multirow{2}{*}{T5}     & Strict  & 0.4382        & 0.5431      & 0.4851 \\
                        & Approx. & 0.4820        & 0.7100      & 0.5634 \\
\hline
\end{tabular}}
\end{table}

Table~\ref{tab:t5-entity-linking} shows the entity-linking evaluation scores on the test set for the T5 model.
We applied BERT-based entity-linking to T5 NER out puts for our submission because the text-to-text T5-based scores were lower. Even though the text-to-text based entity-linking did not perform as well as the BERT-based entity-linking, we believe it has strong potential for improvement.Unlike our BERT-based normalization where only word embeddings are used, the text-to-text based entity-linking naturally takes the entire sentence as a context for the normalization.

\begin{table}[htbp]
\caption{\label{tab:text-to-text-indexing}Precision, recall, and F1-score of the chemical indexing on the test set using T5 and GPT models.}
\centering
\resizebox{.8\columnwidth}{!}{
\begin{tabular}{llccc}
\hline
Model &         & \textbf{Prec} & \textbf{Rec}&\textbf{F1} \\ \hline
\multirow{2}{*}{T5}     & Strict  & 0.0302        & 0.6972      & 0.0578 \\
                        & Approx. & 0.0679        & 0.6383      & 0.1187 \\
\hline
\multirow{2}{*}{GPT}    & Strict  &   0.0588   &    0.0392   & 0.0471 \\
                        & Approx. &  0.2337  &  0.1338     & 0.1433 \\
\hline
\end{tabular}}
\end{table}

We attempt the chemical indexing task by providing the title, two sentences from the abstract, and keywords as context and prompting T5 and GPT to generate index terms. Due to the maximum sequence length limit on model inputs, we can only feed two abstract sentences at a time into the models. We repeatedly combine consecutive abstract sentence pairs with the corresponding article title and key words until the model has been prompted with, and made predictions for, the entire article abstract. 

Since the training dataset provided is very small, we train on a ~17 million article subset of PubMed with their respective chemical indexings as labels. 
Table~\ref{tab:text-to-text-indexing} shows evaluation results of the chemical indexing task using T5 and GPT models. They are low, but we expect they can be improved by training on a larger subset or even on the entire PubMed dataset with chemical topics indexed.
Other experiments to try include using models with longer sequence length, or using a variation on the iterative approach recently developed for long document summarization ~\cite{wu2021recursively}.

\section{Official Submission Evaluation Scores}

We have submitted the results on NER and entity-linking using BERT-based and T5-based text-to-text methods for official evaluation.
The official evaluation scores for the NER (Chemical Mention Recognition) are shown in Table~\ref{tab:official_ner}.
Table~\ref{tab:official_linking} shows the official entity-linking (Chemical Normalization to MeSH IDs) (Chemical Normalization to MeSH IDs) evaluation scores on the recognized chemical mentions.

\begin{table}[h]
    \centering
    \caption{Official submission NER (Chemical Mention Recognition) evaluation scores.}
    \resizebox{1\columnwidth}{!}{
    \begin{tabular}{@{}lrrrrrr@{}}
    \toprule
    & \multicolumn{3}{c}{Strict} & \multicolumn{3}{c}{Approximate} \\ \cmidrule(lr){2-4} \cmidrule(lr){5-7}
    File & Precision & Recall & F1 & Precision & Recall & F1 \\ 
    \midrule
    BERT-based & 0.854 & 0.860 & 0.857 & 0.927 & 0.924 &  0.925  \\
    Text-to-Text & 0.782 & 0.555 & 0.649 & 0.854 & 0.599 & 0.704 \\
    \bottomrule
    \end{tabular}}
    \vspace{12pt}
    \label{tab:official_ner}
\end{table}

\begin{table}[h]
    \centering
    \caption{Official submission entity-linking (Chemical Normalization to MeSH IDs) evaluation scores on the recognized chemical mentions.}
    \resizebox{1\columnwidth}{!}{
    \begin{tabular}{@{}lrrrrrr@{}}
    \toprule
    & \multicolumn{3}{c}{Strict} & \multicolumn{3}{c}{Approximate} \\ \cmidrule(lr){2-4} \cmidrule(lr){5-7}
    File & Precision & Recall & F1 & Precision & Recall & F1 \\ 
    \midrule
    BERT-based & 0.433 & 0.654 & 0.521 & 0.442 & 0.811 &  0.566  \\
    Text-to-Text & 0.439 & 0.539 & 0.484 & 0.484 & 0.722 & 0.574 \\
    \bottomrule
    \end{tabular}}
    \vspace{12pt}
    \label{tab:official_linking}
\end{table}

\section{Discussion}

Consistent with previous findings, we achieve good performance by fine tuning larger pre-trained BERT-based language models with in-domain vocabulary sets on the NER task. The performance was further enhanced using a model ensemble. We use the best ensemble for one of our two final NER submissions. Our other NER submission came from text-to-text predictions made by our T5 model. 

As part of this challenge, we also experimented with novel text-to-text based methods using T5 and GPT generative models. We obtained decent overall results, but they are still lower than the results from our best BERT-based models. A key advantage of the text-to-text approach, often also called ``prompting'' methods, is that we are able to perform different sub-tasks without modifying the model architecture.
While we do not observe state-of-the-art results as in some recent literature ~\cite{brown2020language,wei2021finetuned}, these studies suggest that model size needs to be at least 5 billion parameters to have comparable performance with supervised methods. We hypothesize our relatively smaller model size is a factor that affected the results from our text-to-text experiments. Another possible issue is the domain mismatch between the general domain text used to pretrain these models and the biomedical text on which we deployed them. We hypothesize that better results can be obtained by pretraining these generative language models on in-domain text.

There are many other factors that can affect the performance of the text-to-text based approach using generative models. For example, hyper-parameters for text generation and post-processing such as beam-search can greatly affect the final performance. Choices about pretraining and fine-tuning methods are also important factors. Finally, the form of the prompts themselves can have a non-trivial impact on performance. 

We achieve very good performance using BERT-based supervised methods, but the new text-to-text based methods have great potential. We look forward to closing the gap with continuing research in this area.

\printbibliography

@article{devlin2018bert,
  title={Bert: Pre-training of deep bidirectional transformers for language understanding},
  author={Devlin, Jacob and Chang, Ming-Wei and Lee, Kenton and Toutanova, Kristina},
  journal={arXiv preprint arXiv:1810.04805},
  year={2018}
}

@article{lee2020biobert,
  title={BioBERT: a pre-trained biomedical language representation model for biomedical text mining},
  author={Lee, Jinhyuk and Yoon, Wonjin and Kim, Sungdong and Kim, Donghyeon and Kim, Sunkyu and So, Chan Ho and Kang, Jaewoo},
  journal={Bioinformatics},
  volume={36},
  number={4},
  pages={1234--1240},
  year={2020},
  publisher={Oxford University Press}
}

@article{beltagy2019scibert,
  title={Scibert: A pretrained language model for scientific text},
  author={Beltagy, Iz and Lo, Kyle and Cohan, Arman},
  journal={arXiv preprint arXiv:1903.10676},
  year={2019}
}

@article{gu2020domain,
  title={Domain-specific language model pretraining for biomedical natural language processing},
  author={Gu, Yu and Tinn, Robert and Cheng, Hao and Lucas, Michael and Usuyama, Naoto and Liu, Xiaodong and Naumann, Tristan and Gao, Jianfeng and Poon, Hoifung},
  journal={arXiv preprint arXiv:2007.15779},
  year={2020}
}

@article{shin2020biomegatron,
  title={BioMegatron: Larger biomedical domain language model},
  author={Shin, Hoo-Chang and Zhang, Yang and Bakhturina, Evelina and Puri, Raul and Patwary, Mostofa and Shoeybi, Mohammad and Mani, Raghav},
  journal={arXiv preprint arXiv:2010.06060},
  year={2020}
}

@article{brown2020language,
  title={Language models are few-shot learners},
  author={Brown, Tom B and Mann, Benjamin and Ryder, Nick and Subbiah, Melanie and Kaplan, Jared and Dhariwal, Prafulla and Neelakantan, Arvind and Shyam, Pranav and Sastry, Girish and Askell, Amanda and others},
  journal={arXiv preprint arXiv:2005.14165},
  year={2020}
}

@article{raffel2019exploring,
  title={Exploring the limits of transfer learning with a unified text-to-text transformer},
  author={Raffel, Colin and Shazeer, Noam and Roberts, Adam and Lee, Katherine and Narang, Sharan and Matena, Michael and Zhou, Yanqi and Li, Wei and Liu, Peter J},
  journal={arXiv preprint arXiv:1910.10683},
  year={2019}
}

@article{radford2019language,
  title={Language models are unsupervised multitask learners},
  author={Radford, Alec and Wu, Jeffrey and Child, Rewon and Luan, David and Amodei, Dario and Sutskever, Ilya and others},
  journal={OpenAI blog},
  volume={1},
  number={8},
  pages={9},
  year={2019}
}

@article{wu2021recursively,
  title={Recursively Summarizing Books with Human Feedback},
  author={Wu, Jeff and Ouyang, Long and Ziegler, Daniel M and Stiennon, Nissan and Lowe, Ryan and Leike, Jan and Christiano, Paul},
  journal={arXiv preprint arXiv:2109.10862},
  year={2021}
}

@article{wei2021finetuned,
  title={Finetuned Language Models Are Zero-Shot Learners},
  author={Wei, Jason and Bosma, Maarten and Zhao, Vincent Y and Guu, Kelvin and Yu, Adams Wei and Lester, Brian and Du, Nan and Dai, Andrew M and Le, Quoc V},
  journal={arXiv preprint arXiv:2109.01652},
  year={2021}
}

@inproceedings{liu-etal-2021-self,
    title = "Self-Alignment Pretraining for Biomedical Entity Representations",
    author = "Liu, Fangyu  and
      Shareghi, Ehsan  and
      Meng, Zaiqiao  and
      Basaldella, Marco  and
      Collier, Nigel",
    booktitle = "Proceedings of the 2021 Conference of the North American Chapter of the Association for Computational Linguistics: Human Language Technologies",
    month = jun,
    year = "2021",
    address = "Online",
    publisher = "Association for Computational Linguistics",
    url = "https://aclanthology.org/2021.naacl-main.334",
    doi = "10.18653/v1/2021.naacl-main.334",
    pages = "4228--4238",
}

@article{sevgili2021neural,
      title={Neural Entity Linking: A Survey of Models Based on Deep Learning}, 
      author={Ozge Sevgili and Artem Shelmanov and Mikhail Arkhipov and Alexander Panchenko and Chris Biemann},
      year={2021},
      eprint={2006.00575},
      archivePrefix={arXiv},
      primaryClass={cs.CL}
}

@article{bc7track2,
  title={Overview of the NLM-Chem BioCreative VII track: Full-text Chemical Identification and Indexing in PubMed articles},
  author={Robert Leaman, Rezarta Islamaj and Zhiyong Lu},
  journal={Proceedings of the seventh BioCreative challenge evaluation workshop},
  year={2021}
}

@article{bc7track2data,
  title={The chemical corpus of the NLM-Chem BioCreative VII track: Full-text Chemical Identification and Indexing in PubMed articles},
  author={Rezarta Islamaj and Robert Leaman and David Cissel and Meng Cheng and Cathleen Coss and Joseph Denicola and Carol Fisher and Rob Guzman and Preeti Kochar and Nicholas Miliaras and Zoe Punske and Keiko Sekiya and Dorothy Trinh and Deborah Whitman and Susan Schmidt and Zhiyong Lu},
  journal={Proceedings of the seventh BioCreative challenge evaluation workshop},
  year={2021}
}

@article{lipscomb2000medical,
  title={Medical subject headings (MeSH)},
  author={Lipscomb, Carolyn E},
  journal={Bulletin of the Medical Library Association},
  volume={88},
  number={3},
  pages={265},
  year={2000},
  publisher={Medical Library Association}
}

@misc{drugprot,
  title={Overview of DrugProt BioCreative VII track: quality evaluation and large scale text mining of drug-gene/protein relations},
  author={Miranda, Antonio and Mehryary, Farrokh and Luoma, Jouni and Pyysalo, Sampo and Valencia, Alfonso and Krallinger, Martin},
  journal={Proceedings of the seventh BioCreative challenge evaluation workshop},
  year={2021}
}

@article{krallinger2015chemdner,
  title={The CHEMDNER corpus of chemicals and drugs and its annotation principles},
  author={Krallinger, Martin and Rabal, Obdulia and Leitner, Florian and Vazquez, Miguel and Salgado, David and Lu, Zhiyong and Leaman, Robert and Lu, Yanan and Ji, Donghong and Lowe, Daniel M and others},
  journal={Journal of cheminformatics},
  volume={7},
  number={1},
  pages={1--17},
  year={2015},
  publisher={BioMed Central}
}

@article{leitner2010overview,
  title={An overview of BioCreative II. 5},
  author={Leitner, Florian and Mardis, Scott A and Krallinger, Martin and Cesareni, Gianni and Hirschman, Lynette A and Valencia, Alfonso},
  journal={IEEE/ACM Transactions on Computational Biology and Bioinformatics},
  volume={7},
  number={3},
  pages={385--399},
  year={2010},
  publisher={IEEE}
}

\end{document}